\documentclass[runningheads]{llncs}

\usepackage{eccv}

\usepackage{eccvabbrv}

\usepackage{graphicx}
\usepackage{booktabs}

\usepackage{multicol,multirow}
\usepackage{graphicx}
\usepackage{float}

\usepackage[accsupp]{axessibility}  

\usepackage{hyperref}

\usepackage{orcidlink}

\begin{document}

\title{ViFeEdit: A Video-Free Tuner of Your Video Diffusion Transformer} 

\titlerunning{ViFeEdit: A Video-Free Tuner of Your Video Diffusion Transformer}

\author{Ruonan Yu\inst{1}, 
Zhenxiong Tan\inst{1},
Zigeng Chen\inst{1},
Songhua Liu\inst{2}\thanks{Corresponding authors.},
Xinchao Wang\inst{1}\protect\footnotemark[1]}

\authorrunning{R.~Yu et al.}

\institute{National University of Singapore \and
Shanghai Jiao Tong University \\
\email{\{ruonan, zhenxiong, zigeng99\}@u.nus.edu, liusonghua@sjtu.edu.cn, xinchao@nus.edu.sg}}

\maketitle

\begin{figure}[H]
  \centering
   \includegraphics[width=\linewidth]{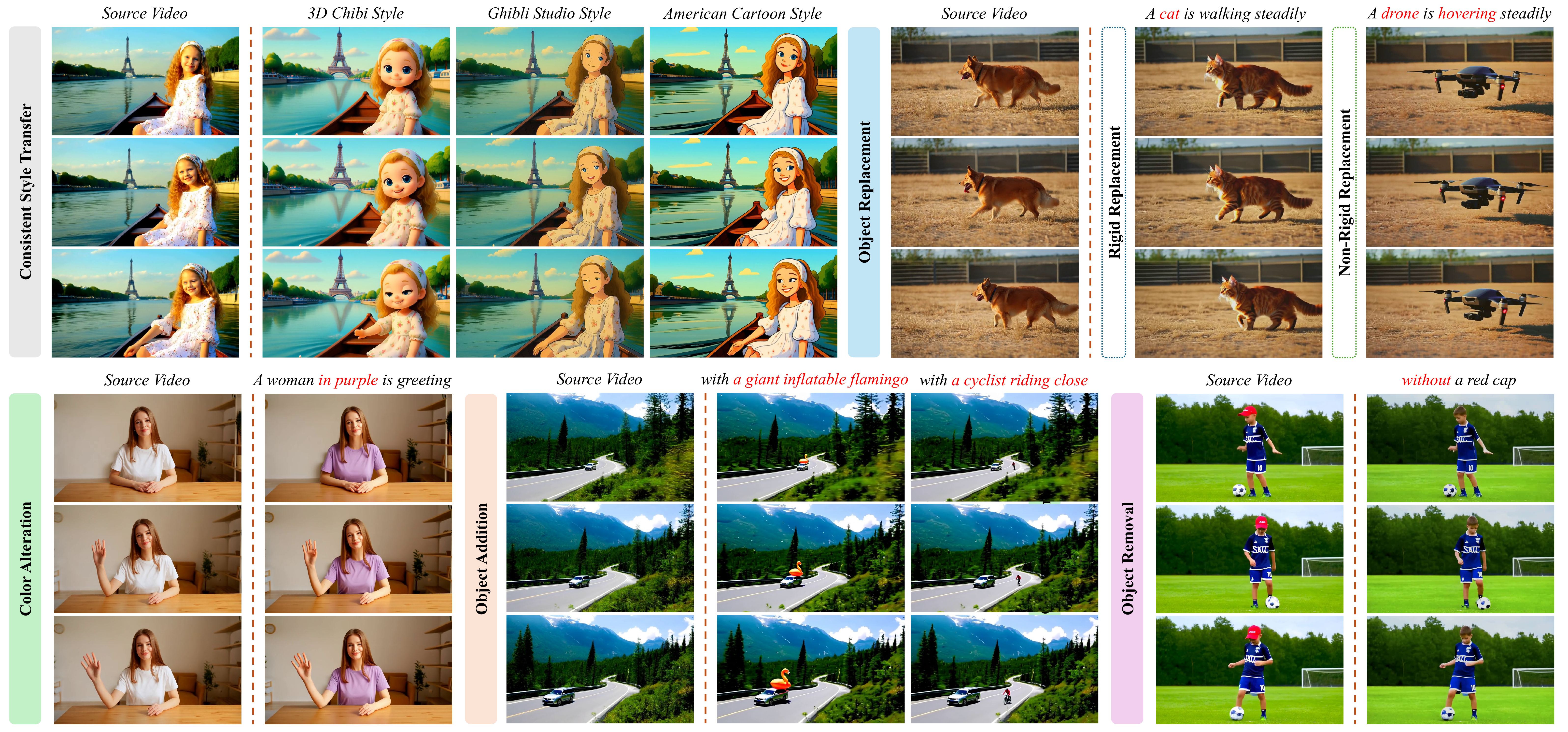}
   \caption{The visualization results of our proposed method ViFeEdit. Our proposed method can adapt text-to-video DiTs to various video editing tasks without any video data. Here, we demonstrate our proposed method on six fine-grained video editing tasks, including style transfer, rigid replacement, non-rigid replacement, color alternation, object addition and object removal.}
   \label{fig:teaser}
\end{figure}

\begin{abstract}
  Diffusion Transformers (DiTs) have demonstrated remarkable scalability and quality in image and video generation, prompting growing interest in extending them to controllable generation and editing tasks. However, compared to the image counterparts, progress in video control and editing remains limited, mainly due to the scarcity of paired video data and the high computational cost of training video diffusion models. To address this issue, in this paper, we propose a video-free tuning framework termed \textbf{ViFeEdit} for video diffusion transformers. Without requiring any forms of video training data, ViFeEdit achieves versatile video generation and editing, adapted solely with 2D images. At the core of our approach is an architectural reparameterization that decouples spatial independence from the full 3D attention in modern video diffusion transformers, which enables visually faithful editing while maintaining temporal consistency with only minimal additional parameters. Moreover, this design operates in a dual-path pipeline with separate timestep embeddings for noise scheduling, exhibiting strong adaptability to diverse conditioning signals. Extensive experiments demonstrate that our method delivers promising results of controllable video generation and editing with only minimal training on 2D image data. Codes are available \href{https://github.com/Lexie-YU/ViFeEdit}{here}. 
\end{abstract}

\section{Introduction}
Diffusion transformers (DiTs)~\cite{ho2020denoising, peebles2023scalable,esser2024scaling} have recently emerged as a highly effective backbone for both image \cite{rombach2022high,podell2023sdxl,zhang2023text,xie2024sana,xie2025sana} and video generation \cite{chen2025sana,zheng2024open,lin2024open,yang2024cogvideox,huang2025self,wan2025wan,kong2024hunyuanvideo,blattmann2023stable,ma2025step}, exhibiting strong generative quality and favorable scaling behavior. Recently, to further effectively accommodate the growing diversity of user requirements, attention has been devoted to controllable generation and editing tasks \cite{zhang2023adding,zhang2025scaling,ruiz2023dreambooth}. 

By training on large-scale paired datasets, current DiTs~\cite{tan2025ominicontrol,tan2025ominicontrol2,zhang2025easycontrol,labs2025flux,wu2025qwen,huang2025diffusion,brooks2023instructpix2pix} have achieved high fidelity and strong usability in image control and editing tasks. 
However, its counterpart in video control and editing~\cite{xing2024survey,sun2024diffusion,hu2023videocontrolnet,jiang2025vace,ma2025controllable,zi2025minimax,gao2025lora,bian2025videopainter,yang2025videograin,yu2025veggie,chen2025perception,huang2025dive} is substantially more challenging since it requires not only spatially coherent modifications, as in image editing, but also their temporally consistent propagation, achieving joint spatiotemporal coherence. 
Moreover, constructing paired video datasets~\cite{yuan2025opens2v,wang2023internvid,bai2025scaling} is substantially more demanding than for images, owing to the increased temporal complexity and the expensive frame-level annotation required for temporal alignment.
For instance, recent efforts~\cite{bai2025scaling} to curate such datasets reportedly consumed over 10,000 GPU days. 
Even with these datasets available, training models for effective video editing and control remains highly resource-intensive due to the inherent multi-frame dependency of video data, typically feasible only for industrial laboratories equipped with large-scale GPU clusters. 


Motivated by these drawbacks, we are curious about one question: \emph{Can a DiT for video editing be effectively tuned without videos, using only 2D images?} 
In this paper, we answer this question affirmatively by introducing \emph{ViFeEdit}, a video-free tuner for video diffusion transformers that enables DiT-based video editors to perform diverse video control and editing tasks with minimal training cost. 
At the core of \emph{ViFeEdit} lies a structural decoupling of spatial and temporal modeling within a DiT-based video generator. 
Specifically, we disentangle the spatial token modeling from the temporal dimension, allowing the tuner to learn spatial editing behaviors purely from 2D images.
Meanwhile, the pretrained temporal modules of the base video generator remain intact, preserving its inherent capability to maintain temporal coherence across frames. 
This design enables \emph{ViFeEdit} to adapt to various video editing tasks without compromising temporal consistency or requiring any video-based supervision. 

However, achieving a clean spatiotemporal decoupling in state-of-the-art DiT architectures, such as the Wan series, is highly non-trivial, as they typically adopt a 3D full-attention mechanism that jointly models spatial and temporal tokens in a unified interaction space. 
As a result, it is difficult to specify which parts of the computation correspond to spatial reasoning and which to temporal reasoning. 
Recent studies~\cite{xi2025sparse} further reveal that modern DiTs dynamically allocate spatial or temporal attention heads depending on the input prompt and diffusion timestep, which makes the decoupling problem even more challenging. 

In this paper, we propose an architectural reparameterization technique to address the above challenge.
Instead of explicitly enforcing a hard separation within the original 3D attention, we introduce a pair of mutually complementary 2D spatial attention blocks that are dedicated to spatial modeling.
On the one hand, these two blocks are initialized to counteract each other, which enables sign-aware semantic editing through decoupled enhancement and suppression in spatial attention. Also, it allows the model to reuse the rich spatial priors of pre-trained 3D attention layers and preserve its original behavior at initialization and thus providing a stable starting point for adaptation.
On the other hand, since the original 3D attention components are entirely frozen, the pretrained temporal modeling capability remains untouched.
Consequently, even when the model is trained solely on 2D images, it can still generate temporally stable and coherent videos during inference. 



Moreover, to further enhance performance, we introduce a dual-path pipeline that separately processes latent states and conditional signals. 
By assigning distinct timestep embeddings for noise scheduling to each branch, this design facilitates more stable optimization and faster convergence. 
We conduct extensive experiments in six fine-grained video editing tasks~\cite{huang2024vbench,li2025five}, including style transfer, rigid object replacement, non-rigid object replacement, color alteration, object addition, and object removal, to validate the effectiveness of our method.
Results demonstrate that our approach enables text-to-video diffusion models to perform diverse editing tasks with minimal computational cost, requiring only a limited amount of image data (100–250 pairs). 

Our contributions can be summarized as follows:
\begin{itemize}
    \item To the best of our knowledge, we present the first approach that adapts text-to-video DiTs to diverse video editing tasks in a video-free scheme; 
    \item To preserve temporal consistency, we introduce an architectural reparameterization that decouples spatial interactions from the full 3D attention and operates within a dual-path pipeline using separate timestep embeddings.
    \item Extensive experiments demonstrate that, with only limited image data and minimal computational cost, our proposed method achieves promising performance across a wide spectrum of video editing tasks. 
\end{itemize}

\section{Related Works}
In this section, we summarize recent progress in diffusion-based video editing approaches. These approaches can be broadly categorized into three paradigms: (1) temporal-adaptation method that explicitly incorporate temporal modules into image backbones~\cite{wu2023tune,gao2025lora,huang2025dive,ma2025magicstick,shin2024edit}, (2) training-free plug-and-play attention- and latent-modulation methods~\cite{li2024vidtome,lu2024fuse,wang2023zero,qi2023fatezero,khachatryan2023text2video,shen2025qk,yang2025videograin,geyer2023tokenflow} that manipulate attention or latent representations during inference, and (3) end-to-end video editing methods~\cite{cheng2023consistent,jiang2025vace,zi2025minimax,yu2025veggie,ye2025unic,ju2025editverse,bai2025scaling} that train a video-conditioned generative model on paired or synthetic supervision to directly produce edited videos.

\textbf{Temporal-Adaptation Methods.} 
These approaches~\cite{wu2023tune,gao2025lora,huang2025dive,ma2025magicstick,shin2024edit} extend pre-trained image diffusion models by explicitly incorporating temporal modeling to ensure cross-frame consistency. They typically inject temporal modules or recurrent connections into pre-trained image models to capture motion dynamics and temporal representations. While effective in improving temporal coherence, such pipelines are computationally expensive and typically require additional training or per-video fine-tuning to learn motion dynamics, which may limit their scalability in real-world applications.

\textbf{Attention- and Latent-Modulation Methods.} 
To enhance efficiency, attention- or latent-based strategies~\cite{li2024vidtome,lu2024fuse,wang2023zero,qi2023fatezero,khachatryan2023text2video,shen2025qk,yang2025videograin,geyer2023tokenflow} modulate spatial or temporal attention within existing diffusion architectures. By reusing frozen image backbones without full temporal training, these approaches achieve higher efficiency lower memory cost. However, their editing capacity is largely confined to appearance-level refinements, making them insufficient for structural or large-scale transformations that require deeper spatiotemporal understanding.

\textbf{End-to-End Methods.} 
More recently, a new class of high-capacity video editing frameworks \cite{cheng2023consistent,jiang2025vace,zi2025minimax,yu2025veggie,ye2025unic,ju2025editverse,bai2025scaling} has emerged, trained in a fully supervised manner on large-scale paired video datasets. These models demonstrate impressive editing strength and robust generalization through joint optimization of content and motion. Nevertheless, such performance comes at the expense of massive computational and data requirements, as curating large-scale paired datasets remains both costly and time-consuming.

\section{Method}
\begin{figure}[t]
  \centering
   \includegraphics[width=\linewidth]{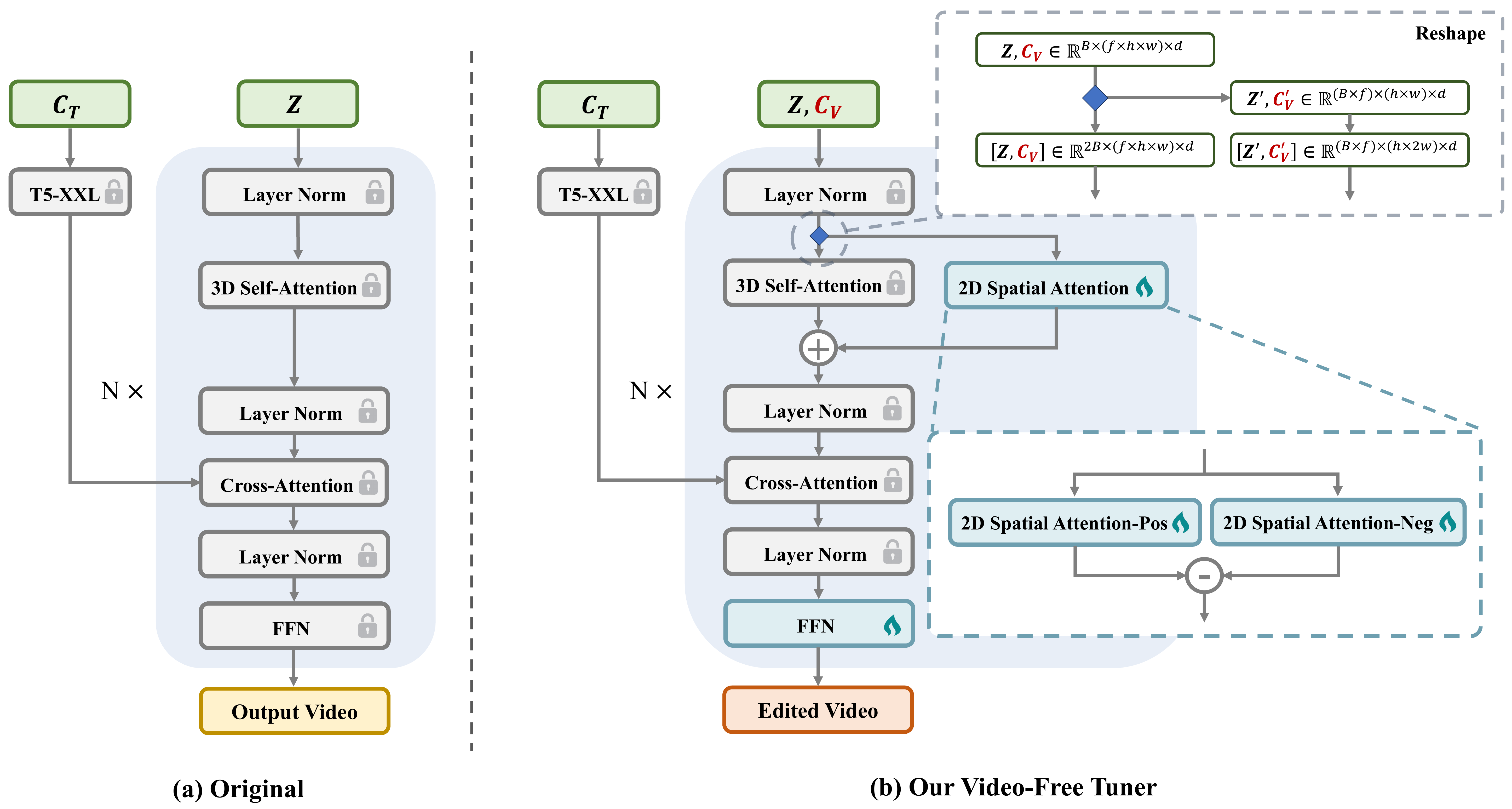}
   \caption{The architecture of DiT blocks of (a) the original text-to-video Wan2.1 model and (b) our proposed video-free tuner ViFeEdit for video editing and control tasks. Here, we enable text-to-video DiTs to handle diverse video editing and control tasks without any video data. Specifically, the source video $C_V$ is jointly fed into the model and interacts with the noisy video latent $Z$ in the 2D spatial attention branch, providing explicit reference guidance.}
   \label{fig:intro}
\end{figure}

In this section, we present the technical details of the proposed \emph{ViFeEdit}. 
We first introduce the preliminaries of DiT-based text-to-video generators in Sec.~\ref{sec:3-1}.
Next, Sec.~\ref{sec:3-3} details the architectural reparameterization technique for spatio-temporal decoupling, which serves as the core of our video-free adaptation framework.
Finally, Sec.~\ref{sec:3-2} elaborates on the dual-path pipeline that enables text-to-video DiTs to perform video editing, including the interaction between the two branches and the separate temporal embedding scheme. 
The overall framework of our proposed method is shown in Fig.~\ref{fig:intro}.

\subsection{Preliminary}\label{sec:3-1}
Effectively capturing spatial and temporal dependencies in the latent space, DiTs~\cite{peebles2023scalable} have been widely adopted in modern video generators, \textit{e.g.}, Wan~\cite{wan2025wan}. 
Typically, a text-to-video DiT takes noisy video latent maps $Z \in \mathbb{R}^{B\times N\times d}$ and text tokens $C_T \in \mathbb{R}^{B\times M\times c}$ as inputs.
Here, $B$ denotes the batch size, $N$ and $M$ represent the number of video and text tokens, respectively, while $d$ and $c$ denote the feature dimensions of the video and text embeddings.
In particular, $N = f \times h \times w$, where $f$ is the number of frames and $h$ and $w$ are the spatial dimensions. 
 
To achieve coherent and temporally consistent video generation, modern DiT-based video generators adopt full 3D attention to jointly capture spatial and temporal dependencies for smooth and stable video results.
\begin{equation}
\begin{aligned}
    \mathrm{Attn}_{3D}(X)=\mathrm{Attention}&(XW^Q,XW^K,XW^V)W^O,\\
    \mathrm{Attention}(Q,K,V)&=\mathrm{Softmax}(\frac{QK^\top}{\sqrt{d'}})V,
\end{aligned}
\end{equation}
where $X$ denotes the hidden state at a given DiT layer, $W^Q$, $W^K$, $W^V$, and $W^O$ are the learnable parameters, and $d'$ represents the dimensionality of this attention feature space. 

During training, they apply the Flow Matching mechanism \cite{lipman2022flow} and obtain noisy video latent maps $Z_t$ with $t\in[0,1]$ by:
\begin{equation}
    Z_t=t\epsilon+(1-t)Z_0,\ v_t=\epsilon-Z_0,\ \epsilon\sim \mathcal{N}(0,I). 
\end{equation}
The parameters $\theta$ of the DiT function $u$ are optimized using the following objective:
\begin{equation}
    \mathcal{L}=\mathbb{E}_{\epsilon,(Z_0,C_T),t}[\Vert u_\theta(Z_t,C_T,t)-v_t\Vert^2].
\end{equation}
In this paper, with only minimal additional parameters, we adapt the text-to-video DiTs to handle various video editing and control tasks without any video training data. We introduce our proposed video-free tuning framework ViFeEdit in the following sections.

\subsection{Spatio-Temporal Decoupling}\label{sec:3-2}
As shown in Sec.~\ref{sec:4}, directly fine-tuning the full 3D attention using only 2D images can disrupt the temporal dynamics inherent in videos, leading to frozen frames during inference. 
The key to addressing this issue lies in a spatio-temporal decoupling mechanism that enables fine-tuning solely the spatial component with 2D images while preserving the model original temporal patterns. 

Explicit decoupling within the full 3D attention module is incompatible with this setting, as the spatial-temporal roles of attention heads vary across denoising steps and conditioning prompts~\cite{xi2025sparse}. 
We tackle this challenge through an architectural reparameterization technique.
Specifically, we keep the original 3D attention untouched and additionally introduce a pair of complementary 2D spatial attention modules. This positive–negative attention architecture facilitates sign-aware semantic editing, where positive and negative semantic signals are explicitly disentangled to enable controlled enhancement and suppression within spatial attention.
Here, the 2D spatial attention modules are initialized with the parameters of the corresponding 3D attention module to reuse the rich spatial priors of pre-trained 3D attention layers and ensure training stability. Also, these 2D attention modules are designed to interact in a residual manner, such that their combined output is zero at initialization, thereby preserving original performance of the model. 
Formally, the final result incorporating the original 3D attention can be written as:
\begin{equation}
    \mathrm{Attn}_{3D}(X)+\mathrm{Attn}_{SpaPos}(X')-\mathrm{Attn}_{SpaNeg}(X'), 
\end{equation}
where $X'$ represents $X$ with a consistent frame index used for the temporal position embedding across all latent frames, and $\mathrm{Attn}_{SpaPos}$ and $\mathrm{Attn}_{SpaNeg}$ denote the newly introduced spatial attention modules, each operating independently on individual frames and computing attention only within the spatial domain. 

To enable fine-tuning using only 2D images, we only update the positive and negative spatial attention modules, $\mathrm{Attn}_{SpaPos}$ and $\mathrm{Attn}_{SpaNeg}$, as well as the feed-forward layers to enhance performance.
Again, the original 3D attention remains frozen during fine-tuning to preserve its pretrained temporal generation capability.

\subsection{Dual-Path Pipeline}\label{sec:3-3}
Building upon the proposed spatio-temporal decoupling technique, the remaining challenge is to equip the DiT with the ability to take a source video as input and effectively inject conditional information into the backbone features. 
Inspired by recent image editing approaches~\cite{tan2025ominicontrol,tan2025ominicontrol2,zhang2025easycontrol}, instead of introducing a separate encoder, we reuse the DiT backbone to encode the conditional information. 
However, unlike previous approaches that directly concatenate conditional tokens with noisy latent tokens and allow them to interact throughout all attention layers, we adopt a dual-path pipeline.
Specifically, the two streams are processed separately and only interact within the positive and negative spatial attention modules introduced above, ensuring that the original 3D attention remains intact and its temporal generation capability is preserved. 

In other words, the 3D attention treats the noisy video latents $Z\in\mathbb{R}^{B\times N\times d}$, $N=f\times h\times w$, and the video condition $C_V\in\mathbb{R}^{B\times N\times d}$ as independent samples by concatenating them along the batch dimension, \textit{i.e.}, $[Z, C_V] \in \mathbb{R}^{2B\times N\times d}$, and assigning them separate 3D position embeddings as usual. 
For spatial attention modules, we flatten the inputs $Z$ and $C_V$ into $\mathbb{R}^{(B\times f)\times(h\times w)\times d}$ as the single-frame videos with batch size $B\times f$ and concatenated along the spatial dimension, \textit{i.e.}, either $h$ or $w$, ensuring the interaction is  within each frame.
Without loss of generality, when concatenation occurs along the $w$ dimension, we assign positional indices within $[0,2w)$ for this axis, while setting the temporal positional indices to $0$ for all tokens.  
This design enables the model to learn rich editing and control tasks mapping solely from 2D paired image training data, while strengthening the frame-wise consistency between the generated video $Z$ and the original input video $C_V$.

Optionally, in order to further enhance structural consistency, inspired by SDEdit~\cite{meng2021sdedit}, $C_V$ can be used as a noise prior to initialize the noisy latent during inference:
\begin{equation}
Z_\alpha = (1-\alpha) C_V + \alpha \epsilon,
\end{equation}
where $\alpha\in[0,1]$ is a hyper-parameter controlling the strength of the prior. 
The flow-matching schedule then starts from $t = \alpha$. 

\textbf{Separate Timestep Embeddings.} 
During training and inference, $Z$ and $C_V$ correspond to the noisy latent map and the clean source video, respectively. 
As a result, they exhibit distinct noise levels, and using the same timestep input for both can blur the conditional guidance. 
To address this issue, we assign separate timestep embeddings to $Z$ and $C_V$, ensuring reliable conditional injection during both training and inference. 
Specifically, for $Z$, the timestep is the current flow-matching timestep $t$ as usual, while for $C_V$, the timestep is always $0$, indicating a clean video input. 
These separate embeddings are concatenated along the batch dimension accordingly. 

\begin{figure}[t]
  \centering
   \includegraphics[width=\linewidth]{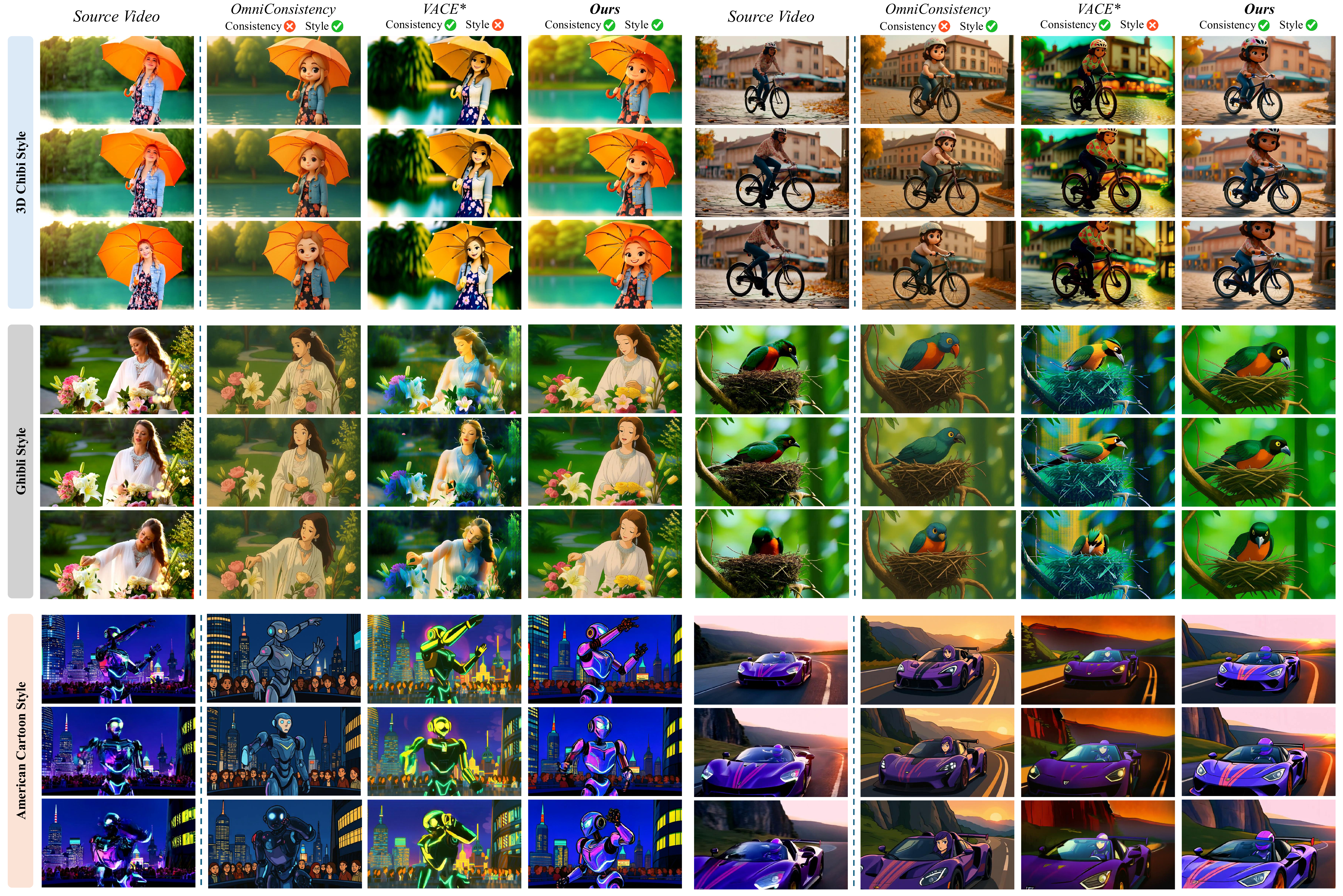}
   \caption{The visualization results of baselines and our proposed method on consistent style transfer tasks. $^*$ means that the pretrained model VACE is further finetuned on the paired image data for style transfer learning.}
   \label{fig:style}
\end{figure}

\begin{table}[t]
\centering
\caption{The evaluation results of baselines and our proposed method on style transfer task. Here, the experiments are conducted on three target styles, 3D Chibi style, Ghibli Studio style, and American Cartoon style. $^*$ means that the pretrained VACE is further finetuned on the paired image data for style transfer learning. $\dagger$ means that the evaluation of consistency is conducted between the source and target video.}
\label{tab:baseline} 
\resizebox{\linewidth}{!}{
\begin{tabular}{cccccccccccccc}
\toprule
\midrule
\multirow{3}{*}{\bf Method/Setting} && \multicolumn{5}{c}{\textbf{VBench}} && \multicolumn{3}{c}{\textbf{VLM Score}} \\
\cmidrule{3-7}\cmidrule{9-11}
 && \textbf{Subject}  & \textbf{Background} & \textbf{Temporal} & \textbf{Motion} & \multirow{2}{*}{\bf Color} && \textbf{Structural} & \textbf{Motion} & \multirow{2}{*}{\bf Stylization}\\
 && \textbf{Consistency} & \textbf{Consistency} & \textbf{Flickering} & \textbf{Smoothness} &&& \textbf{Consistency$^\dagger$} & \textbf{Consistency$^\dagger$} & \\ 
\midrule
\multicolumn{11}{c}{\textbf{\textit{3D Chibi Style}}}\\
\midrule
OmniConsistency && 0.9711 & 0.9712 & 0.9948 & 0.9811 & 0.9113 && 90.86 & 89.67 & \bf 93.06\\
VACE$^*$ && 0.9751 & 0.9758 & 0.9945 & 0.9805 & 0.8737 && 84.68 & 86.29 & 91.13\\
\bf Ours && \bf 0.9811 & \bf 0.9785 & \bf 0.9980 & \bf 0.9872 & \bf 0.9259 && \bf 91.13 & \bf 90.16 & \bf 93.06\\
\midrule
\multicolumn{11}{c}{\textbf{\textit{Ghibli Studio Style}}}\\
\midrule
OmniConsistency && 0.9689 & 0.9715 & 0.9946 & 0.9805 & 0.9023 && 90.80 & 90.48 & 93.06\\
VACE$^*$ && 0.9680 & 0.9709 & 0.9949 & 0.9828 & 0.7521 && 89.84 & 88.87 & 91.77\\
\bf Ours && \bf 0.9773 & \bf 0.9777 & \bf 0.9978 & \bf 0.9866 & \bf 0.9106 && \bf 93.39 & \bf 92.42 & \bf 93.39\\
\midrule
\multicolumn{11}{c}{\textbf{\textit{American Cartoon Style}}}\\
\midrule
OmniConsistency && 0.9712 & 0.9699 & 0.9915 & 0.9702 & \bf 0.9075 && 88.87 & 88.79 & 91.33\\
VACE$^*$ && 0.9718 & 0.9778 & 0.9942 & 0.9797 & 0.8349 && 85.16 & 86.45 & 89.19\\
\bf Ours && \bf 0.9802 & \bf 0.9789 & \bf 0.9974 & \bf 0.9844 & 0.8764 && \bf 91.46 & \bf 90.81 & \bf 91.46\\
\midrule
\bottomrule
\end{tabular}
}
\end{table}

\begin{figure}[t]
  \centering
   \includegraphics[width=\linewidth]{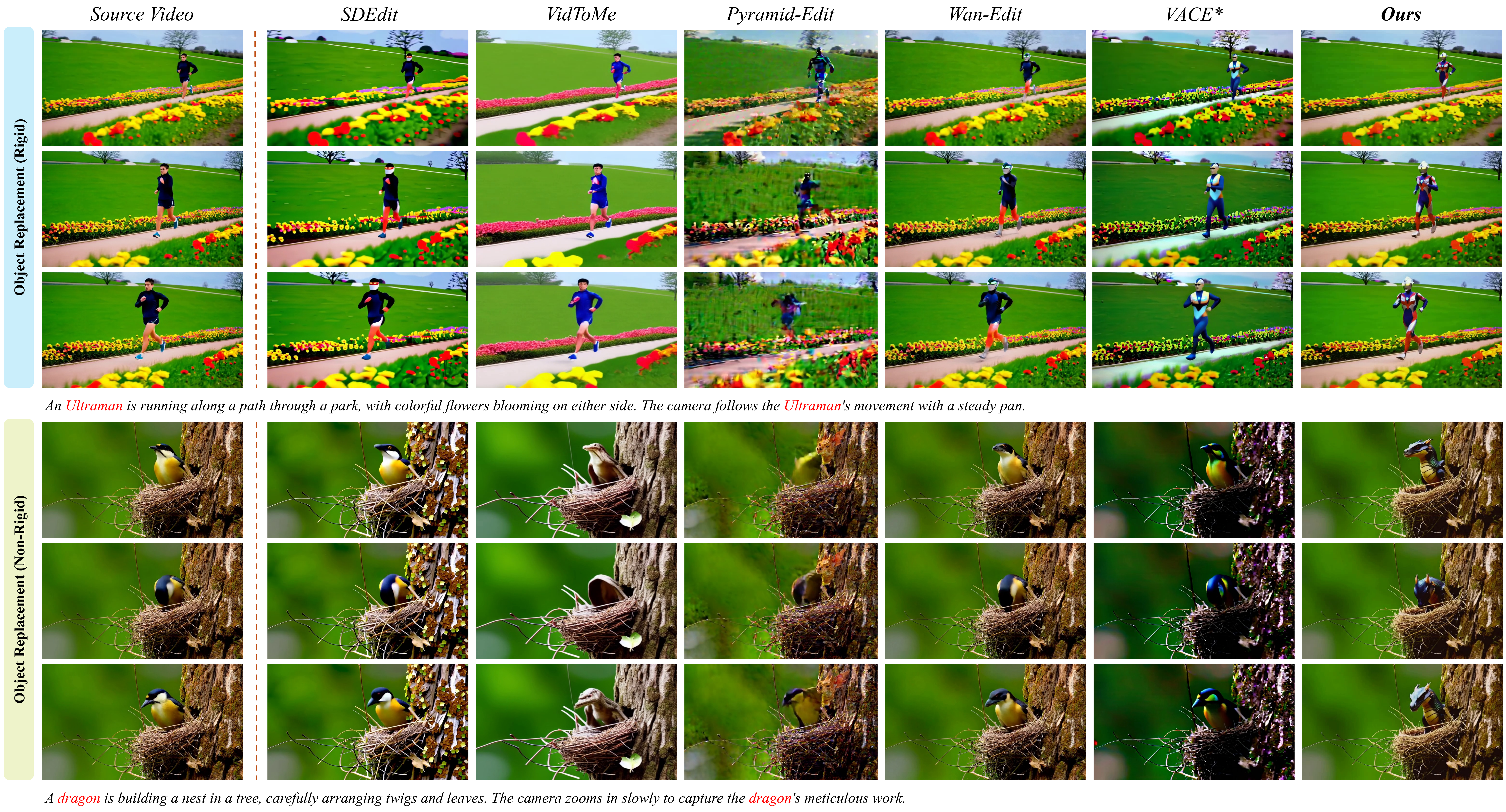}
   \caption{The visualization results of baselines and our proposed method on rigid and non-rigid replacement tasks. $^*$ means that the pretrained VACE is further finetuned on the paired image data for editing tasks.}
   \label{fig:edit}
\end{figure}

\begin{figure}[t]
  \centering
   \includegraphics[width=\linewidth]{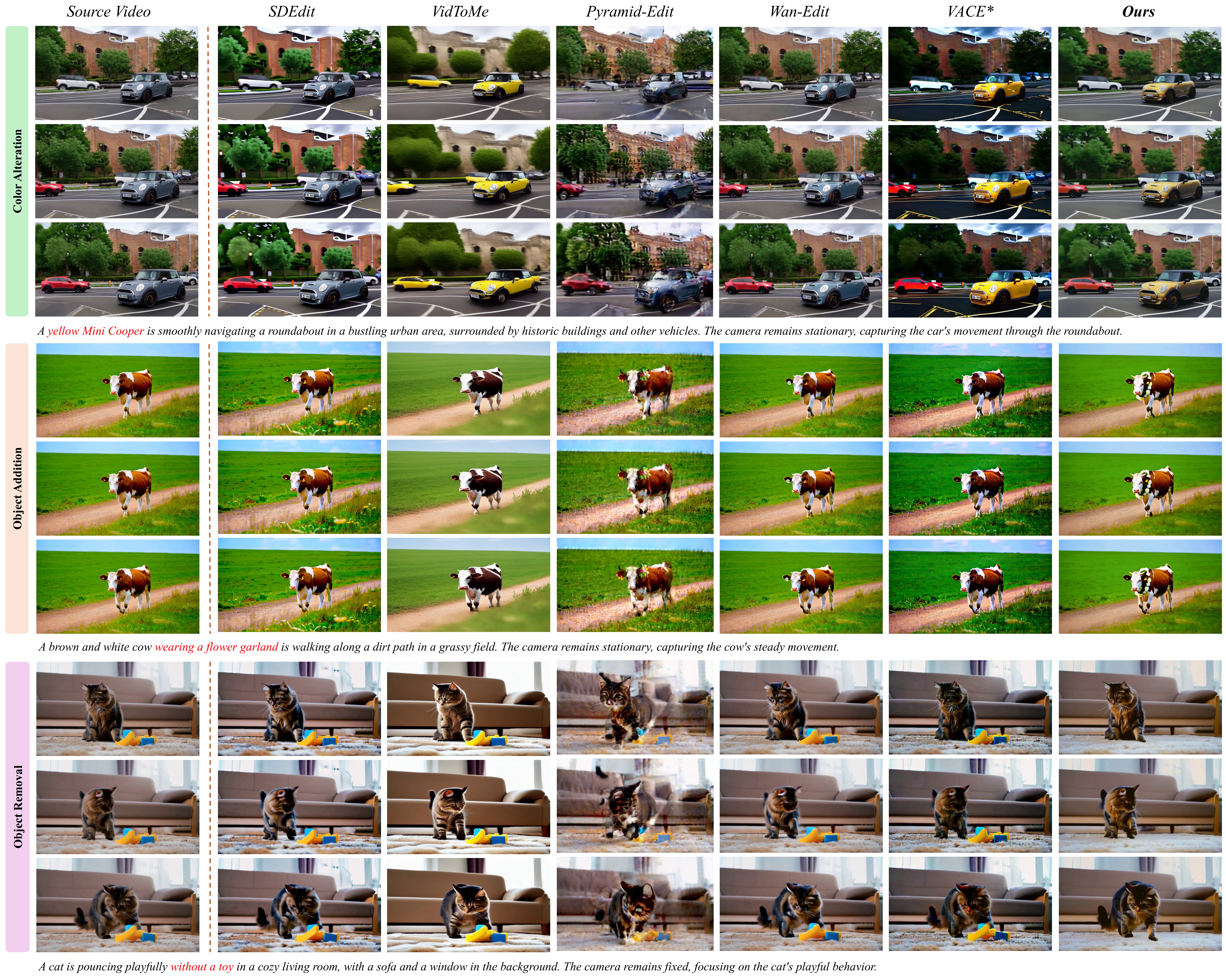}
   \caption{The visualization results of baselines and our proposed method on color alternation, object addition and object removal tasks. $^*$ means that the pretrained VACE is further finetuned on the paired image data for editing tasks.}
   \label{fig:edit2}
\end{figure}

\section{Experiments}\label{sec:4}
\subsection{Settings and Implementation Details}
In this paper, we propose a video-free tuning framework, ViFeEdit, to enable text-to-video diffusion transformers to handle various video editing and control tasks with solely 2D paired image data. To validate the effectiveness of our proposed method, we conduct comprehensive experiments on 6 video editing tasks, $i.e.$, consistent style transfer, rigid object replacement, non-rigid object replacement, color alteration, object addition, and object removal. We also conduct experiments on depth-to-video generation, please refer to the supplementary for more results. Here, we adopt the open-source text-to-video model Wan2.1-T2V-1.3B~\cite{wan2025wan} as the base model. 
\subsubsection{Finetuning Settings and Details}
Here, for the consistent style transfer task, we adopt the open-source image dataset OmniConsistency~\cite{song2025omniconsistency}, which contains 100-200 paired samples for each style. With only the limited paired image data, our method achieves stable and high-quality video stylization results. For the remaining editing tasks, we adopt GPT-5 to randomly generate prompts for editing tasks and then adopt FLUX.1-dev to generate the source images and Qwen-Image-Edit-2509~\cite{wu2025qwen} to generate the corresponding target edited images. Each task consists of 250 paired samples. Each image data is treated as single-frame video. During training, we employ LoRA fine-tuning~\cite{hu2022lora}, which is both efficient and lightweight. The rank is set to 32 for all tasks, and the training typically lasts within 20 epochs, yielding high-quality editing results for all tasks. 
\subsubsection{Evaluation Settings and Details}
For the style transfer task, we follow the official VBench evaluation settings \cite{huang2024vbench}. We generate five base videos for each prompt and apply consistent style transfer methods to obtain stylized videos. The resulting stylized videos are then evaluated using the subject consistency, background consistency, temporal flickering, motion smoothness and color metrics provided by VBench, which collectively measure visual quality, temporal consistency. Further, we evaluate VLM score with Qwen2.5-VL-7B-Instruct~\cite{bai2025qwen2} on structural consistency and motion consistency between the base video and the stylized video, and stylization quality of the target video for style fidelity.
As for other editing tasks, \textit{e.g.}, rigid and non-rigid object replacement, color alteration, object addition, and object removal, we adopt the FiVE-Bench~\cite{li2025five}, following its provided task prompts to generate base videos and perform edit. We evaluate the edited results using the FiVE-Acc metrics, which offers a comprehensive quantitative measure of editing accuracy.
Specifically, to obtain more comprehensive results, the FiVE-Acc metrics are evaluated over entire videos rather than a few sampled frames, ensuring accuracy and stability.

\subsubsection{Baseline Settings and Details}
For the consistent style transfer task, we adopt the powerful end-to-end model Wan2.1-VACE-1.3B~\cite{jiang2025vace}, which is pretrained on large video datasets, as baseline. To enable the VACE model to handle unseen consistent style transfer tasks, we perform LoRA fine-tuning on the vace branch using the same image dataset OmniConsistency \cite{song2025omniconsistency}, with the rank set to 32 for all styles. Moreover, we adopt OmniConsistency method \cite{song2025omniconsistency} to conduct frame-by-frame style transfer on the base video for comparison, and all experiments are conducted following the official settings and checkpoint. Here, all videos are of 81 frames and the resolution is 480p. 
As for other editing tasks, we adopt SDEdit~\cite{meng2021sdedit}, VidToMe~\cite{li2024vidtome}, Pyramid-Edit~\cite{li2025five,jin2024pyramidal}, Wan-Edit~\cite{li2025five,kulikov2025flowedit}, and Wan2.1-VACE-1.3B as baselines. Here, SDEdit and Wan-Edit are based on Wan2.1-T2V-1.3B model and the denoising strength is set to 0.7, following the default settings. VidToMe is based on Stable-Diffusion-1.5, and Pyramid-Edit is based on Pyramid-Flow-miniFLUX model. VACE is LoRA fine-tuned with the same paired image datasets as ours for each editing task to learn editing mappings. Here, all videos are also of 81 frames and 480p resolution, except VidToMe for 63 frames and Pyramid-Edit for 384p followed by official default settings.

\subsection{Results on Consistent Style Transfer}
For the consistent style transfer task, we adopt the Wan2.1-T2V-1.3B model to generate the base videos with prompts provided by VBench. Following the default settings of VBench, we generate five videos for each prompt using seeds 0–4, and then apply style transfer methods to them. For this task, we evaluate the effectiveness of our proposed method on 3D Chibi, Ghibli, and American Cartoon styles. 
The evaluation results are reported in Table~\ref{tab:baseline} and the visualization samples are shown in Fig.~\ref{fig:style}. For more experimental results and visualizations of comparison with Ditto-14B model~\cite{bai2025scaling}, please refer to the supplementary.

As shown in the results, our proposed method can achieve high-quality style transfer with both temporal consistency and spatial consistency. Here, the OmniConsistency method can preserve the transferred style with stable color retention. However, as OmniConsistency is based on the image diffusion model FLUX-dev, and it performs frame-by-frame style transfer, it suffers from poor temporal consistency and struggles to capture coherent motion. Also, when transferring blurred or ambiguous frames, the stylized results often introduces noticeable abrupt changes. This issue can also be seen at the first example of the American Cartoon style in Fig.~\ref{fig:style}, where a sudden shift occurs due to the lack of temporal modeling.
VACE can preserve temporal consistency. However, it suffers from substantial color drift and unstable style adherence, as shown in the second example of Ghibli studio style in Fig.~\ref{fig:style}. It may come from compensation for its limited style knowledge learned from the image data by over-relying on color changes, which leads to visible chromatic fluctuations. It is further supported by the color metrics in Table~\ref{tab:baseline}. With limited paired image data for training, our method achieves high-quality style transfer while simultaneously preserving spatial structure and temporal coherence. It maintains stable color behavior and captures the target style more faithfully, producing the most visually consistent and stylistically accurate results across all evaluated settings.

\begin{table}[t]
\centering
\caption{The evaluation results of baselines and our proposed method on FiVE-Benchmark with object replacement~(rigid and non-rigid), color alternation, object addition, and object removal tasks. $^*$ means that the pretrained VACE is further finetuned on the paired image data for editing tasks.}
\label{tab:edit}
\resizebox{\linewidth}{!}{
\begin{tabular}{ccccccccccccccccccccccccccc}
\toprule
\midrule
\multirow{2}{*}{\bf Method/Setting} && \multicolumn{5}{c}{\textbf{Object Replacement~(Rigid \& Non-Rigid)}} && \multicolumn{5}{c}{\textbf{Color Alteration}} \\
\cmidrule{3-7} \cmidrule{9-13}
&& \bf YN-Acc  & \bf MC-Acc & \bf $\cup$-Acc & \bf $\cap$-Acc & \bf FiVE-Acc && \bf YN-Acc  & \bf MC-Acc & \bf $\cup$-Acc & \bf $\cap$-Acc & \bf FiVE-Acc \\
\midrule
SDEdit && 21.00 & 32.50 & 33.50 & 20.00 & 26.75 && 13.00 & 22.00 & 23.00 & 12.00 & 17.50 \\
VidToMe && 36.50 & 51.00 & 51.50 & 36.00 & 43.75 && 47.00 & 51.00 & 54.00 & 44.00 & 49.00\\
Pyramid-Edit && 48.00 & 67.50 & 68.00 & 47.50 & 57.75 && 52.00 & 49.00 & 59.00 & 42.00 & 50.50\\
Wan-Edit && 39.50 & 52.50 & 53.00 & 39.00 & 46.00 && 37.00 & 41.00 & 44.00 & 34.00 & 39.00\\
VACE$^*$ && 20.50 & 34.00 & 34.00 & 20.50 & 27.25 && 82.00 & 87.00 & 89.00 & 80.00 & 84.50\\
\bf Ours && \bf 72.00 & \bf 83.50 & \bf 84.00 & \bf 71.50 & \bf 77.75 && \bf 87.00 & \bf 96.00 & \bf 98.00 & \bf 85.00 & \bf 91.50\\
\midrule
\multirow{2}{*}{\bf Method/Setting}&& \multicolumn{5}{c}{\textbf{Object Addition}} && \multicolumn{5}{c}{\textbf{Object Removal}}\\
\cmidrule{3-7} \cmidrule{9-13}
&& \bf YN-Acc  & \bf MC-Acc & \bf $\cup$-Acc & \bf $\cap$-Acc & \bf FiVE-Acc && \bf YN-Acc  & \bf MC-Acc & \bf $\cup$-Acc & \bf $\cap$-Acc & \bf FiVE-Acc \\
\midrule
SDEdit && 0.00 & 11.11 & 11.11 & 0.00 & 5.56 && 0.00 & 0.00 & 0.00 & 0.00 & 0.00\\
VidToMe && 11.11 & 11.11 & 11.11 & 11.11 & 11.11 && 0.00 & 10.00 & 10.00 & 0.00 & 5.00\\
Pyramid-Edit && 66.67 & 88.89 & 88.89 & 66.67 & 77.78 && 0.00 & 0.00 & 0.00 & 0.00 & 0.00\\
Wan-Edit && 33.33 & 66.67 & 66.67 & 33.33 & 50.00 && 0.00 & 0.00 & 0.00 & 0.00 & 0.00\\
VACE$^*$ && 22.22 & 22.22 & 22.22 & 22.22 & 22.22 && 0.00 & 0.00 & 0.00 & 0.00 & 0.00\\
\bf Ours && \bf 100.00 & \bf 100.00 & \bf 100.00 & \bf 100.00 & \bf 100.00 && \bf 80.00 & \bf 80.00 & \bf 80.00 & \bf 80.00 & \bf 80.00\\
\midrule
\bottomrule
\end{tabular}
}
\end{table}

\subsection{Results on Video Editing}
We further conduct experiments on a broader set of video editing tasks, \textit{e.g.}, rigid and non-rigid replacement, color alteration, and object addition and removal. Here, we adopt FiVE-Bench as our evaluation benchmark. We compare our method with five video editing methods, SDEdit, VidToMe, Pyramid-Edit, Wan-Edit, and VACE. 
All methods are evaluated using their default settings without extra inputs such as depth maps or masked videos.  
Following the guidelines in FiVE-Bench, we employ the Qwen2.5-VL-7B-Instruct model for evaluation. Notably, our evaluation is performed on the entire video.
The quantitative results are presented in Table~\ref{tab:edit}, with visual examples shown in Fig.~\ref{fig:edit} and Fig.~\ref{fig:edit2}.

The results show that our method achieves effective and high-fidelity edits. For both rigid and non-rigid replacement tasks, VACE$^*$ can do certain editing but introduces background inconsistencies.
SDEdit and Wan-Edit maintain background coherence but often provide  partial edits. 
Pyramid-Edit supports replacement but degrades video quality. 
Our method can perform object replacement with high fidelity in both rigid and non-rigid scenarios, producing detailed target objects that preserve motion and integrate seamlessly with the background. 
For color alteration, transforming dark colors into bright ones is challenging. As shown in Fig.~\ref{fig:edit2}, SDEdit and Wan-Edit only lighten the car, while VidToMe and VACE$^*$ achieves the transformation but VidToMe lacks localized control, unintentionally affecting nearby cars and the background, and VACE$^*$ introduces background color inconsistencies. Our method enables precise color changes without disturbing other regions.
For the fine-grained addition task, SDEdit and Wan-Edit produce only minor local edits and fail to achieve the required modifications. Pyramid-Edit is able to produce the intended additions, but the target video quality is degraded. Our method can perform fine-grained addition edits while preserving the video quality and overall consistency. 
For removal task, as indicated by Table~\ref{tab:edit} and Fig.~\ref{fig:edit2}, baseline methods struggle to fully remove target objects, whereas our method achieves complete removal with plausible background completion.

\begin{table}[t]
\centering
\caption{The performance of multi-task LoRA and single-task LoRA on FiVE-Bench. Multi-task LoRA is trained with objective addition, removal and color alternation tasks at the same time.}
\label{tab:multi}
\resizebox{\linewidth}{!}{
\begin{tabular}{ccccccccccccccccccccccccccc}
\toprule
\midrule
\multirow{2}{*}{\bf Task/Setting} && \multicolumn{5}{c}{\textbf{Single-Task LoRA}} && \multicolumn{5}{c}{\textbf{Multi-Task LoRA}}\\
\cmidrule{3-7} \cmidrule{9-13}
&& \bf YN-Acc  & \bf MC-Acc & \bf $\cup$-Acc & \bf $\cap$-Acc & \bf FiVE-Acc && \bf YN-Acc  & \bf MC-Acc & \bf $\cup$-Acc & \bf $\cap$-Acc & \bf FiVE-Acc \\
\midrule
\bf Color Alternation && 87.00 & 96.00 & 98.00 & 85.00 & 91.50 && 87.00 & 95.00 & 97.00 & 85.00 & 91.00\\
\bf Object Addition && 100.00 & 100.00 & 100.00 & 100.00 & 100.00 && 100.00 & 100.00 & 100.00 & 100.00 & 100.00\\
\bf Object Removal && 80.00 & 80.00 & 80.00 & 80.00 & 80.00 && 80.00 & 90.00 & 90.00 & 80.00 & 85.00\\
\midrule
\bottomrule
\end{tabular}
}
\end{table}

\begin{figure}[t]
  \centering
   \includegraphics[width=\linewidth]{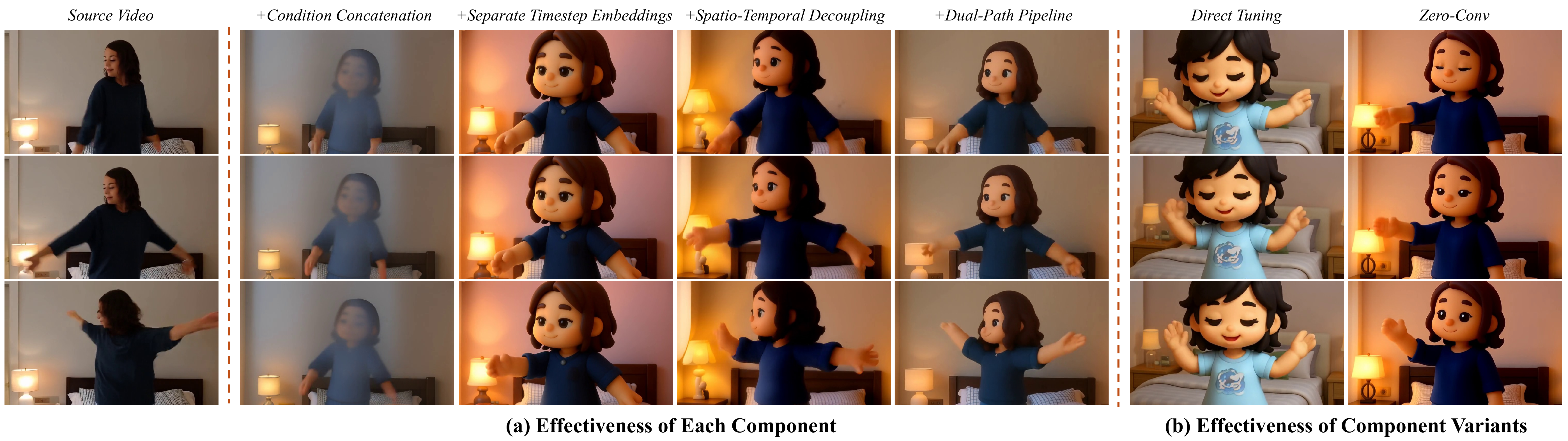}
   \caption{The visualization results of ablation studies. (a) The effectiveness of each proposed component. From left to right, each key component cumulatively builds upon the previous configuration. (b) Comparison with other strategies, including Direct Tuning and Zero-Conv.}
   \label{fig:abl}
\end{figure}

\subsection{Ablation Studies}
\subsubsection{Results on Multi-Task LoRA}
To further demonstrate the scalability and generalization capability of our proposed method, we conduct multi-task LoRA fine-tuning with color alternation, object addition, and object removal tasks at the same time. The results are shown in Table~\ref{tab:multi}. We also conduct multi-style finetuning experiments in Supplementary Section B.2. The results show that our framework does not require training a separate LoRA or model for each editing task or style, and instead supports multiple tasks within a unified framework.
\subsubsection{Effectiveness of Each Key Components}
We further conduct ablation studies on the key components of our method and provide visualizations to more intuitively illustrate the contribution of each component. As shown in Fig.~\ref{fig:abl}(a), we take the 3D Chibi style transfer task as an example.
Here, the starting point is direct tuning with image data and concatenating conditional tokens with noisy tokens. 
From left to right, each step introduces one more key component of our method, cumulatively building upon the previous configuration.
For the starting point, it is difficult for the model to learn clean representations, as the reference video remains noisy during training. To address this problem, we adopt separate timestep embeddings. It assures the clean reference video and helps the model to learn the target style rapidly. Here, the third column only takes less than half the training time of the second column.
However, the motion remains weakened due to the image data training, and the structural consistency is still limited. To address the issue, we further introduce spatio-temporal decoupling. It isolates the learning process on image data to ensure the temporal consistency. To further enhance the spatial consistency, we propose dual-path pipeline. As illustrated in the final column, our proposed method achieves consistent motion, background preservation, and high-quality style transfer.

\subsubsection{Comparison with Other Variants} To further demonstrate the effectiveness of our method, we compare it with several alternative strategies. The visualization results are shown in Fig.~\ref{fig:abl}(b). First, we directly fine-tune the attention layers ($i.e.$, $q$, $k$, $v$, and $o$) as well as the feed-forward layers of the Wan2.1-T2V-1.3B model. As shown in the results, first column of Fig.~\ref{fig:abl}(b), since 3D attention jointly encodes spatial and temporal information, directly finetuning under image-only supervision can disrupt temporal generation capability, leading to degraded motion quality such as frozen frames. This can also be observed in the direct tuning with condition concatenation, as shown in the second column of Fig.~\ref{fig:abl}. Our proposed method introduce 2D spatial attention branch, which decouples spatial-temporal attention from 3D attention and confines image-based updates only to the 2D spatial branch, preserving temporal modeling learned by the text-to-video backbone. Moreover, to further validate the effectiveness of our proposed positive-negative 2D spatial attention, we compare it with ControlNet-style zero-convolution approaches, the results are shown in the second column of Fig.~\ref{fig:abl}(b). Although ControlNet-style zero-convolution can achieve a similar initialization effect, it requires to train additional modules from scratch, which is challenging to converge under image-supervision, resulting in weaker alignment with the source video. Our proposed positive-negative 2D spatial attention reuses the rich spatial priors of pre-trained 3D attention layers, and preserve high consistency.

\section{Conclusion}
In this paper, we propose a video-free tuning framework termed \textit{ViFeEdit} for video diffusion transformers. With solely 2D images, our proposed method can adapt text-to-video DiTs to diverse video editing tasks at minimal costs. Specifically, we propose an architectural reparameterization to decouple the spatial interactions from the full 3D attention. It can isolate the image tuning process and strengthen the visual control ability, without compromising the temporal consistency. Moreover, to further enhance background and motion consistency, we introduce a dual-path pipeline with separate timestep embeddings. It also benefits training process for more stable optimization and faster convergence. Extensive experiments demonstrate that, with limited paired image data, our proposed method enables video diffusion models to achieve promising performance with a wide range video editing tasks.


%
%
\bibliographystyle{splncs04}
\bibliography{main}

\newpage
\appendix
\section{More Experimental Details}
\subsection{More Dataset Details}
For the consistent style transfer task, we use OmniConsistency~\cite{song2025omniconsistency} as the training dataset. We select three style subsets, 3D Chibi, Ghibli, and American Cartoon. These subsets contain 140, 100, and 124 pairs of original and styled images, specifically.
For the rest editing tasks, we use GPT-5 to generate 250 editing prompts for each editing task. For each prompt, we adopt FLUX.1-dev to generate the source image, and Qwen-Image-Edit-2509~\cite{wu2025qwen} to produce the target image according to the specified edit instruction.

For evaluation, we use VBench~\cite{huang2024vbench} prompts and generate five source videos per prompt with Wan2.1-T2V-1.3B~\cite{wan2025wan}, using seeds 0–4. For the resting editing tasks, the source videos are produced by mixing Wan2.1-T2V-1.3B~\cite{wan2025wan} and Wan2.1-T2V-14B~\cite{wan2025wan} with prompts from FiVE Bench~\cite{li2025five}, and the seeds are randomly drawn from 0–4. The number of evaluation samples is 100, 100, 100, 9, 10 for object rigid replacement, object non-rigid replacement, color alternation, object addition, and object removal tasks, respectively.

\subsection{More Model Details}
In this paper, we adopt Wan2.1-T2V-1.3B~\cite{wan2025wan} as the base model, and introduce a pair of complementary 2D spatial attention module applied to each 3D full attention block in residual manner. Both 2D spatial attention blocks are initialized with the corresponding 3D attention module to ensure training stability, and their outputs are combined by subtraction.

To establish strong baselines, we adopt the end-to-end model Wan2.1-VACE-1.3B~\cite{jiang2025vace}, which is pretrained on a large-scale video dataset and provides strong controllability and editing performance. To enable VACE to handle unseen consistent style transfer tasks, we apply LoRA fine-tuning on the VACE branch using the corresponding style subsets of the OmniConsistency dataset~\cite{song2025omniconsistency}, with the LoRA rank fixed to 32 for all styles. We also include the OmniConsistency method applied to the base video frame-by-frame for comparison. OmniConsistency method is based on FLUX.1-dev model. Here, we directly use the corresponding LoRA modules provided officially for each specific style, and the rank for each LoRA module is 128. Here, all source and generated target videos contain 81 frames with 480p resolution.

For other editing tasks, we use SDEdit~\cite{meng2021sdedit}, VidToMe~\cite{li2024vidtome}, Pyramid-Edit~\cite{jin2024pyramidal,li2025five}, and Wan-Edit~\cite{kulikov2025flowedit,li2025five} as baselines for comparison. Here, SDEdit~\cite{meng2021sdedit} and Wan-Edit~\cite{kulikov2025flowedit,li2025five} are based on the Wan2.1-T2V-1.3B model, and we use a denoising strength of 0.7 following the default configuration stated in FiVE Bench. VidToMe~\cite{li2024vidtome} is based on Stable-Diffusion-1.5 model~\cite{rombach2022high}, while Pyramid-Edit~\cite{jin2024pyramidal,li2025five} is based on the Pyramid-Flow-miniFLUX model. All methods generate videos with 81 frames at 480p resolution unless otherwise specified. VidToMe~\cite{li2024vidtome} outputs 63 frames, and Pyramid-Edit~\cite{jin2024pyramidal,li2025five} produces 384p videos, both following their official default settings.

\subsection{More Training Details}
All tasks are fine-tuned with LoRA under the DiffSynth framework. Except for the number of training epochs, all other training hyperparameters are kept consistent across tasks. Specifically, the LoRA rank is set to 32, and only the 2D spatial attention modules and FFN modules are fine-tuned, with the rest part of network remains frozen. Training is performed on three NVIDIA RTX 6000 Ada GPUs. The peak GPU memory is about 18 GiB, and each epoch will take 5 min for consistent style transfer task and 9 min for the rest editing tasks. More implementation details are provided in Table~\ref{tab:hype}.

\begin{table}[ht]
\centering
\caption{The hyper-parameters of training for all tasks. Here, the training epoch from left to right is for consistent style transfer, object rigid replacement, object non-rigid replacement, color alternation, object addition and object removal tasks.}
\label{tab:hype} 
\begin{tabular}{ccc}
\toprule
\bf Hyperparameter && \bf Value \\
\midrule
Optimizer && AdamW\\
Learning Rate && 1e-4 \\
Weight Decay && 0.01\\
LoRA Rank && 32\\
Scheduler && ConstantLR\\
\multirow{3}{*}{LoRA Target Modules} && SpaPos.q, k, v, o\\
&& SpaNeg.q, k, v, o\\
&& ffn.0, ffn.2\\
CFG Scale && 1.0 \\
Training Epoch && 20, 10, 20, 10, 5, 2\\
\bottomrule
\end{tabular}
\end{table}

\begin{figure}[t]
  \centering
   \includegraphics[width=\linewidth]{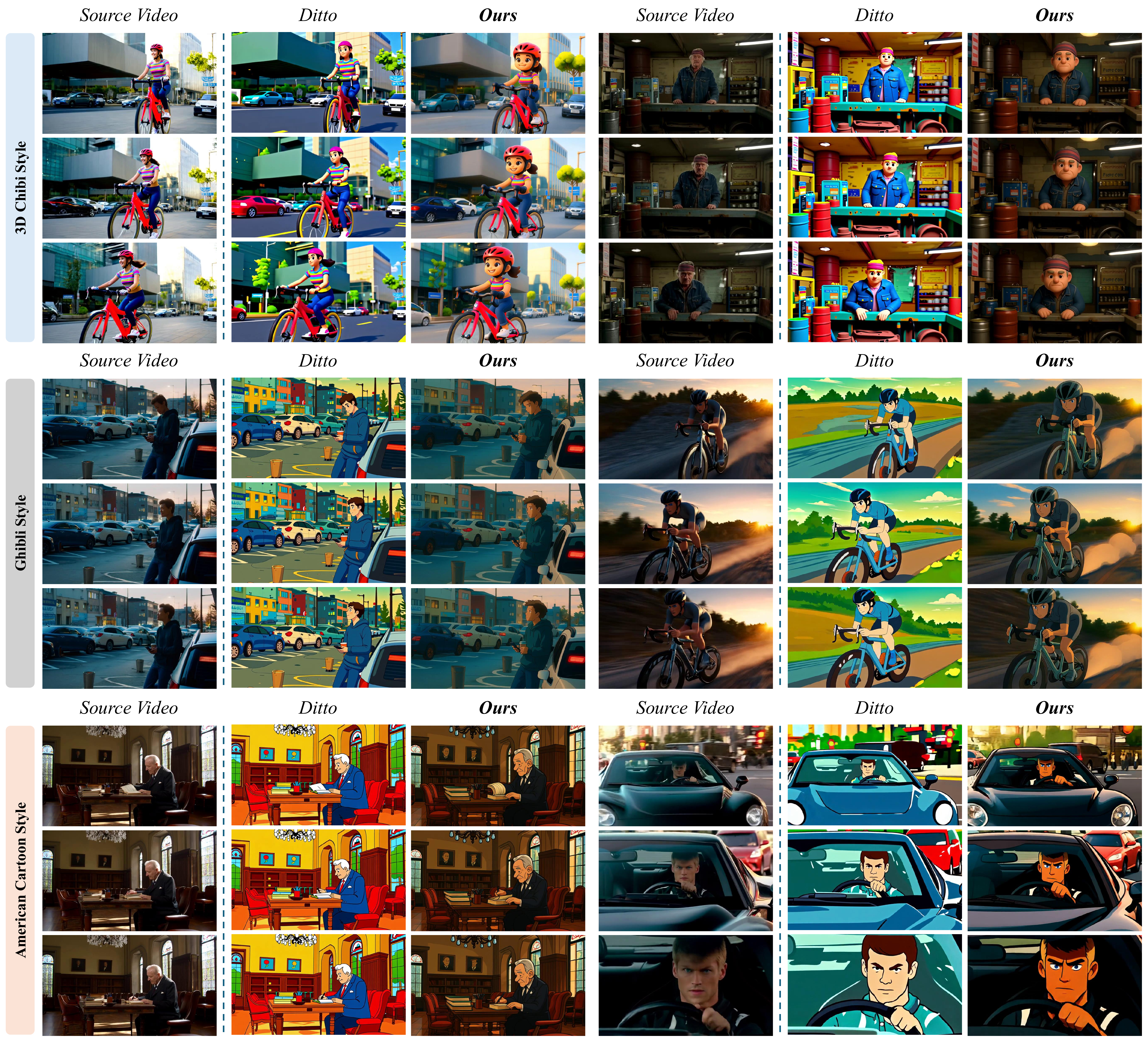}
   \caption{The visualization results of Ditto and our proposed method on consistent style transfer tasks. Here, our proposed method achieves high consistency and style fidelity with only paired image data.}
   \label{fig:ditto}
\end{figure}

\begin{table}[ht]
\centering
\caption{The evaluation results of Ditto and our proposed method on VBench with consistent style transfer task. Here, Ditto has 14B model size and pre-trained with large-scale paired video dataset, while our proposed method is 1.3B and trained on 100-200 paired image data.}
\label{tab:ditto} 
\resizebox{\linewidth}{!}{
\begin{tabular}{cccccccccccccccccccccccccc}
\toprule
\midrule
\multirow{3}{*}{\bf Method/Setting} &\multirow{3}{*}{\bf Model Size} & \multirow{3}{*}{\bf Training Sets} && \multicolumn{5}{c}{\bf VBench} \\
 \cmidrule{5-9}
&&&& \bf Subject & \bf Background & \bf Temporal & \bf Motion & \multirow{2}{*}{\bf Color}\\
&&&& \bf Consistency & \bf Consistency & \bf Flickering & \bf Smoothness\\
\midrule
\multicolumn{9}{c}{\textbf{\textit{3D Chibi Style}}}\\
\midrule
Ditto & 14B & Ditto-1M~(Video) && 0.9749 & 0.9790 & 0.9965 & 0.9759 & 0.7948\\
\bf Ours & 1.3B & OmniConsistency~(Image) && \bf 0.9855 & \bf 0.9806 & \bf 0.9981 & \bf 0.9907 & \bf 0.9357\\
\midrule
\multicolumn{9}{c}{\textbf{\textit{Ghibli Studio Style}}}\\
\midrule
Ditto & 14B & Ditto-1M~(Video) && 0.9823 & \bf 0.9803 & 0.9958 & 0.9853 & 0.7903\\
\bf Ours & 1.3B & OmniConsistency~(Image) && \bf 0.9828 & 0.9802 & \bf 0.9979 & \bf 0.9900 & \bf 0.9054\\
\midrule
\multicolumn{9}{c}{\textbf{\textit{American Cartoon Style}}}\\
\midrule
Ditto & 14B & Ditto-1M~(Video) && 0.9831 & \bf 0.9820 & 0.9952 & 0.9829 & 0.6755\\
\bf Ours & 1.3B & OmniConsistency~(Image) && \bf 0.9850 & 0.9819 & \bf 0.9974 & \bf 0.9884 & \bf 0.8720\\
\midrule
\bottomrule
\end{tabular}
}
\end{table}

\subsection{More Evaluation Details}
For evaluation, we adopt established benchmarks to assess performance across a variety of video editing tasks. All test videos consist of 81 frames. For consistent style transfer task, we follow the official VBench protocol~\cite{huang2024vbench}. We generate five base videos for each prompt, and apply the style-transfer methods produce stylized outputs. We report the subset of VBench metrics that appropriately capture style-transfer behavior, including subject consistency, background consistency, temporal flickering, motion smoothness, and color.
We further compute VLM-based scores using Qwen2.5-VL-7B-Instruct~\cite{bai2025qwen2} to measure structural and motion consistency between the base and stylized videos, as well as style fidelity in the final outputs.
For VLM evaluation, we use the following prompts to evaluate structural and motion consistency, and style fidelity:
for structural consistency, we use: `Score the spatial and structural consistency between the two videos with \textit{prompt}. Give a single score from 0 to 100, where 0 means completely inconsistent and 100 means perfectly consistent. Output only the number, with no other text'; for motion consistency, we use: `Score the temporal and motion consistency between the two videos with \textit{prompt}. Give a single score from 0 to 100, where 0 means completely inconsistent and 100 means perfectly consistent. Output only the number, with no other text'; for style fidelity, we use: `Evaluate the style transfer quality between the two videos with \textit{prompt}. The target style is \textit{style}. Give a single score from 0 to 100, where 0 means the style is completely failed and 100 means the style is perfectly transferred while maintaining visual coherence. Output only the number, with no other text'. Here, we set the $fps$ as 15 to help model thoroughly review the input source and target videos.

For other editing tasks such as rigid and non-rigid object replacement, color modification, object addition, and object removal, we adopt FiVE-Bench and use its task prompts to produce base videos and corresponding edits. Editing performance is quantified using the FiVE-Acc metrics. To improve the reliability of this benchmark, we review and refine the question set used by FiVE-Bench, and we compute FiVE-Acc over entire videos instead of a small set of sampled frames, which may provide more stable and representative quantitative results.

\section{More Experimental Results}
\subsection{Comparison with End-to-End Video Editing Model Ditto}
To further validate the effectiveness of our proposed method, we compare it against Ditto~\cite{bai2025scaling} on the consistent style transfer task. Ditto is built on Wan2.1-VACE-14B~\cite{jiang2025vace} and trained on 1M high-quality video pairs. We use the officially released ditto\_global\_style LoRA, which yields its strongest style-transfer performance. Following the VBench protocol~\cite{huang2024vbench}, we generate one video per prompt. Quantitative results are reported in Table~\ref{tab:ditto}, and visual comparisons are in Fig.~\ref{fig:ditto}.

Both the quantitative results and the visualization results demonstrate that our proposed method, requiring only 100-200 image pairs for fine-tuning, has high consistency and style fidelity style transfer performance across different styles. From the results, Ditto~\cite{bai2025scaling} preserves motion coherence and temporal stability but often fails to maintain accurate style, color correspondence, and spatial alignment. For example, in the second American Cartoon example, Ditto introduces clear color shifts in the car~(from black to blue), the hair of the man~(from yellow to brown), and the clothes of the man~(from black to blue). This failure also reflected by the color metric in Table~\ref{tab:ditto}.

\begin{figure}[t]
  \centering
   \includegraphics[width=\linewidth]{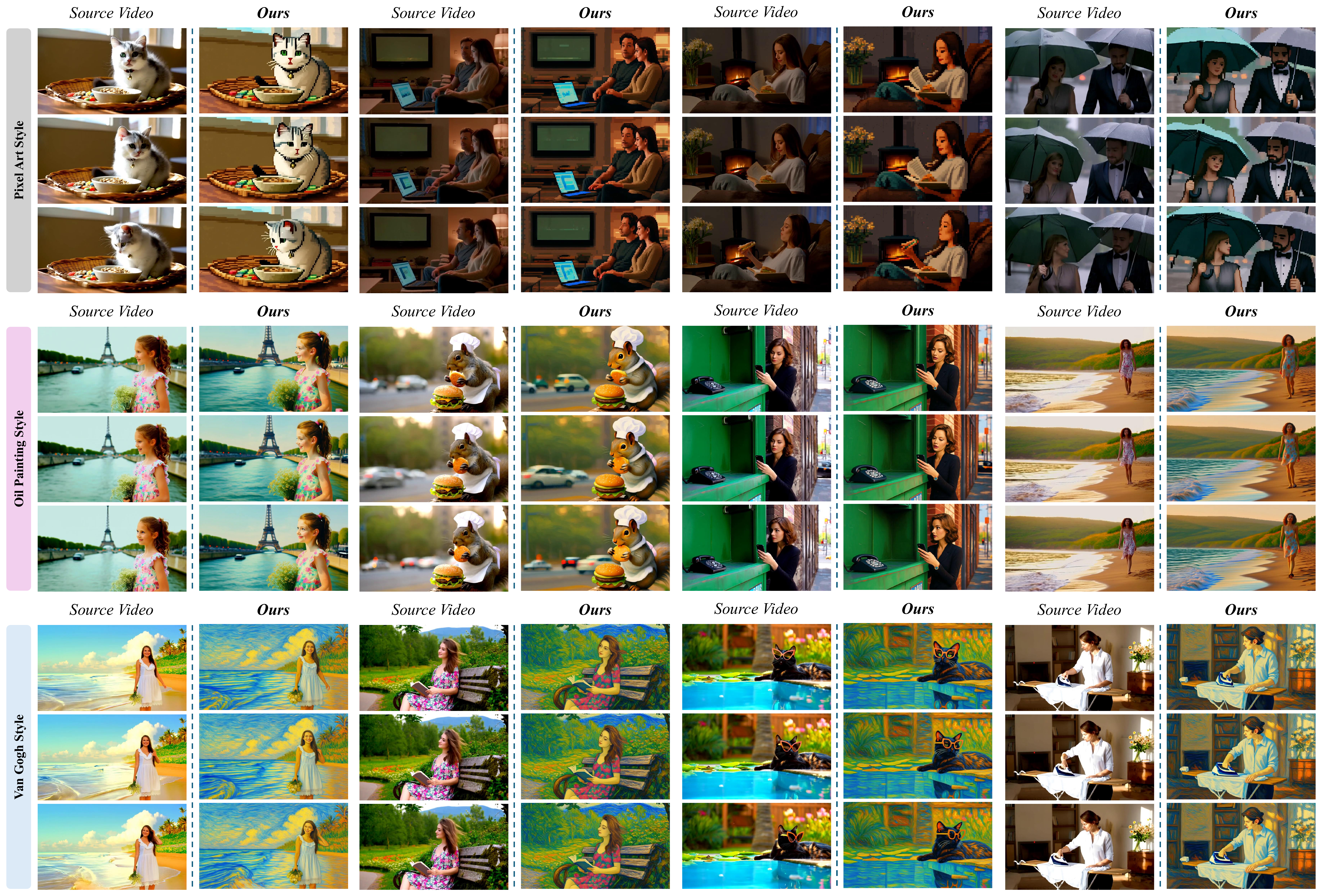}
   \caption{The visualization results of our proposed method on mixed style transfer task. Here, we fine-tune Wan2.1-T2V-1.3B model with mixed style image data. Our proposed method can handle multiple styles transfer at one time.}
   \label{fig:arbitrary}
\end{figure}

\begin{figure}[t]
  \centering
   \includegraphics[width=\linewidth]{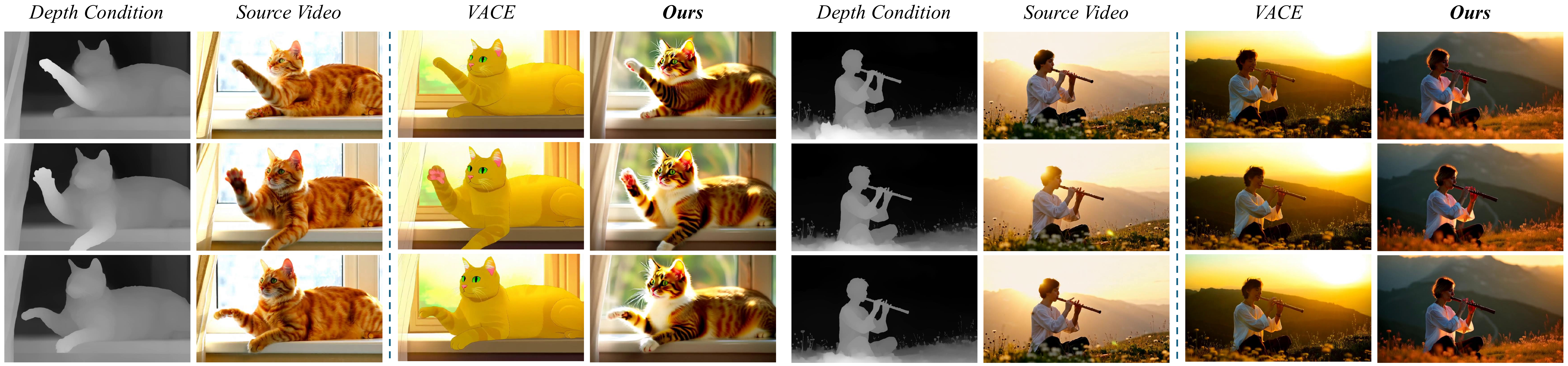}
   \caption{The visualization results of Wan2.1-VACE-1.3B and our proposed method on depth-to-video tasks. Here, the source videos are generated by Wan2.2-T2V-A14B, and the depth maps are generated by Video Depth Anything.}
   \label{fig:depth}
\end{figure}

\subsection{Results on Arbitrary Style Transfer}
In previous experiments, we focus on transferring a single style. To further demonstrate the scalability and generalization capability of our proposed method, we conduct multi-style mixed fine-tuning with style Pixel Art, Oil Painting, and Van Gogh. Here, we combine the subsets with those three styles from OmniConsistency dataset~\cite{song2025omniconsistency}. The total number of paired image data is 317. We fine-tune the Wan2.1-T2V-1.3B model~\cite{wan2025wan} following the settings stated in Table~\ref{tab:hype}. The total fine-tuning stage lasts for 20 epochs. The visualization results are shown in Fig.~\ref{fig:arbitrary}. 

From the results, our method is able to handle multiple styles jointly while maintaining high consistency in both spatial and temporal dimensions. Notably, our proposed method performs high-quality style transformation and can also reliably distinguish and reproduce even closely related styles, such as Van Gogh and classical oil painting, without introducing cross-style interference.

\subsection{Results on Depth-to-Video Task}
To further assess the applicability of our proposed method and its capability to handle different types of control signals, we perform additional experiments on controllable video generation tasks. Here, we conduct experiments on the depth-to-video task. 

Specifically, we first use GPT-5 to randomly generate 250 prompts, and then employ FLUX.1-dev to produce the corresponding images as the target. These targets are paired with depth maps obtained from Depth-Anything-V2~\cite{yang2024depth}, forming the supervision signals required for depth-conditioned video generation. For fine-tuning, we follows the hyper-parameter settings demonstrated in Table~\ref{tab:hype}, and the total number of epochs for fine-tuning is 20. The visualization results are shown in Fig.~\ref{fig:depth}. Here, the source videos are generated by Wan2.2-T2V-A14B~\cite{wan2025wan}, and the corresponding depth maps are generated by Video Depth Anything~\cite{chen2025video}. The prompts of the source video are randomly selected from VBench~\cite{huang2024vbench}. All source videos have 81 frames and are in 480p resolution. Here, to validate the effectiveness of our proposed method, we employ Wan2.1-VACE-1.3B model~\cite{jiang2025vace} as the baseline for comparison. Wan2.1-VACE-1.3B~\cite{jiang2025vace} is pre-trained on large-scale paired video dataset, and has outstanding performance in controllable video generation tasks.

From the results, our proposed method closely adheres to the input depth conditions, capturing fine-grained structural cues with high precision. In addition, the aesthetic quality and overall visual fidelity of the generated videos are comparable to those produced by the Wan2.1-VACE-1.3B model~\cite{jiang2025vace}, which is trained on large-scale video datasets.

\begin{figure}[t]
  \centering
   \includegraphics[width=\linewidth]{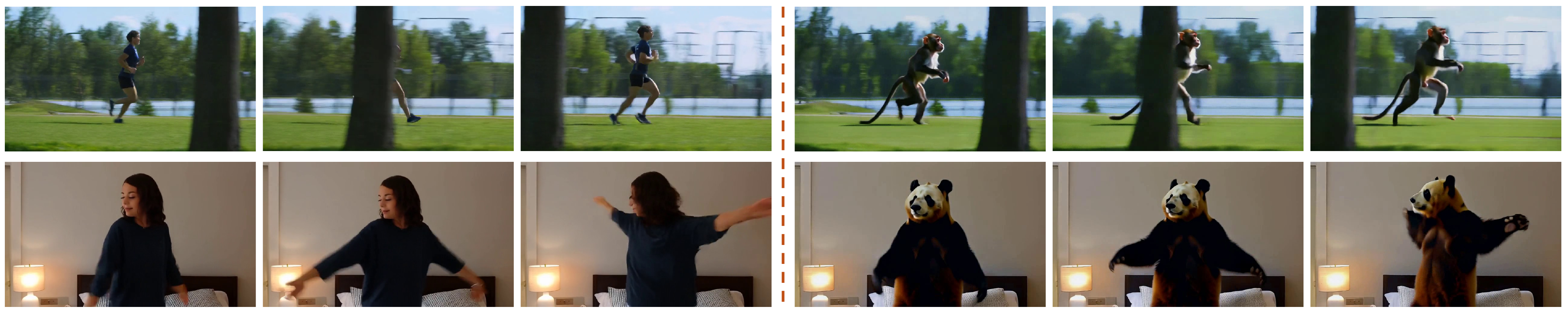}
   \caption{Visualization results of our method on more challenging editing cases. The left three columns are the source videos, and the right columns are edited videos via our proposed method.}
   \label{fig:complex}
\end{figure}

\subsection{Results on Challenging Cases}
To further demonstrate the effectiveness of our proposed method, we apply ViFeEdit on more challenging cases, \textit{e.g.}, motion blur and significant occlusion. The edited results are shown in Fig.~\ref{fig:complex}. From the results, our proposed method performs well and still keeps coherent temporal consistency and stable structural preservation under challenging cases.

\section{Conclusion}
Our method opens a new direction for the video editing and control tasks, and it achieves high-quality and highly consistent performance while using only a small amount of easily obtainable paired image, without requiring large-scale paired video data and extensive training. The experimental results further show that our proposed method attains performance on par with, and in many cases exceeding, large-scale models trained on massive amounts of paired video data with heavy training budgets. Moreover, our proposed method exhibits strong versatility, supporting a wide range of tasks, from global editing to local editing, and even controllable generation, while consistently delivering high-quality outputs.

\end{document}